\pdfoutput=1

\documentclass[11pt]{article}

\usepackage[]{ACL2023}

\usepackage{times}
\usepackage{latexsym}

\usepackage[T1]{fontenc}

\usepackage[utf8]{inputenc}

\usepackage{microtype}

\usepackage{inconsolata}

\usepackage{xcolor}
\usepackage{natbib}
\usepackage{subcaption}
\usepackage{amsmath}
\usepackage{xfrac}
\usepackage[]{hyperref}
\usepackage{booktabs}
\usepackage{multirow}
\usepackage{float}
\usepackage{bbm}

\newcommand{\ColorBestLoss}{\color[HTML]{5C70D4}}
\newcommand{\ColorBestAcc}{\color[HTML]{CC665C}}
\newcommand{\ANBN}{$a^nb^n$}
\newcommand{\REPOURL}{\url{https://github.com/0xnurl/mdl-lstm}}

\title{Bridging the Empirical-Theoretical Gap in Neural Network Formal Language Learning Using Minimum Description Length}

\author{
Nur Lan\textsuperscript{1,2}, Emmanuel Chemla\textsuperscript{1,3}, Roni Katzir\textsuperscript{2}
\\\textsuperscript{1}Ecole Normale Supérieure, \textsuperscript{2}Tel Aviv University, \textsuperscript{3}EHESS, PSL University, CNRS 
\\ 
\texttt{\{nur.lan,emmanuel.chemla\}@ens.psl.eu}\\
\texttt{rkatzir@tauex.tau.ac.il}}

\begin{document}
\maketitle

\begin{abstract}

Neural networks offer good approximation to many tasks but consistently fail to reach perfect generalization, even when theoretical work shows that such perfect solutions can be expressed by certain architectures.
Using the task of formal language learning, we focus on one simple formal language and show that the theoretically correct solution is in fact not an optimum of commonly used objectives --- even with regularization techniques that according to common wisdom should lead to simple weights and good generalization (L1,~L2) or other meta-heuristics (early-stopping, dropout). On the other hand, replacing standard targets with the Minimum Description Length objective (MDL) results in the correct solution \linebreak being an optimum.

\end{abstract}

\begin{figure*}[h!]
    \centering

    \begin{subfigure}{0.3\textwidth}
    \centering
    \includegraphics[width=\textwidth]{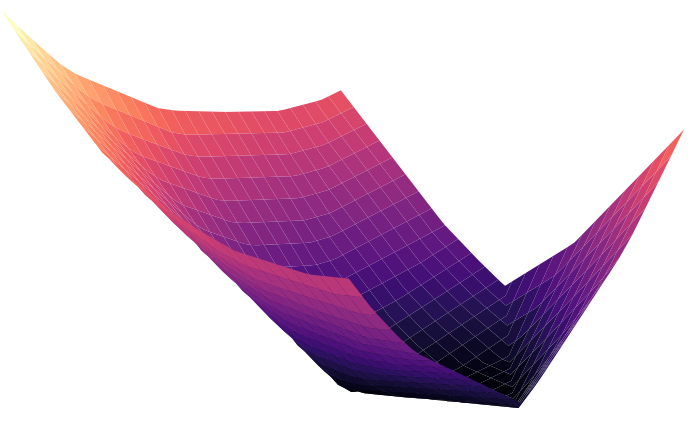}
    \caption{L1}
    \label{fig:regularizations-3d:l1}
    \end{subfigure}\hfill
    \begin{subfigure}{0.3\textwidth}
    \centering
    \includegraphics[width=0.98\textwidth]{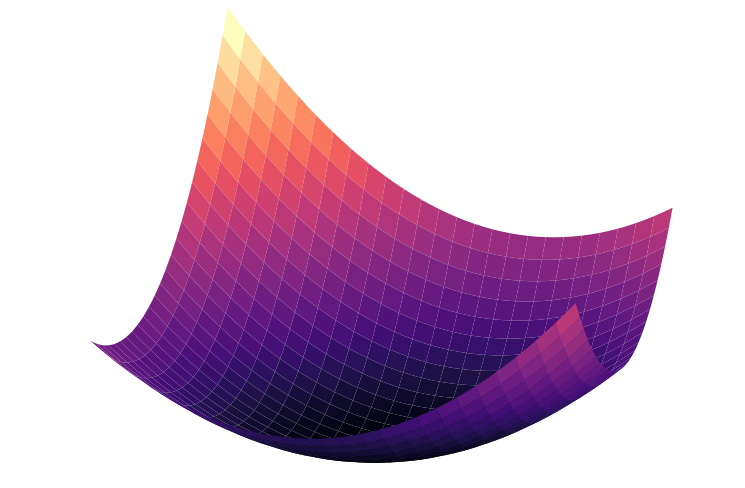}
    \caption{L2}
    \label{fig:regularizations-3d:l2}
    \end{subfigure}\hfill
    \begin{subfigure}{0.3\textwidth}
    \centering
    \includegraphics[width=\textwidth]{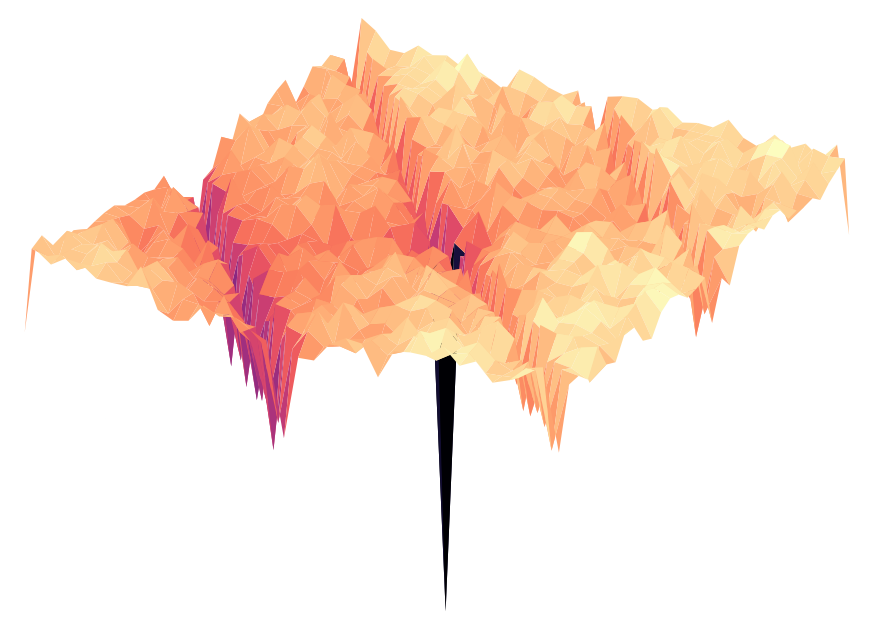}
    \caption{$|H|$}
    \label{fig:regularizations-3d:mdl}
    \end{subfigure}\hfill

\caption{Loss surfaces around the golden $a^nb^n$ LSTM from Section~\ref{sec:experiment-golden-anbn}, for the regularization terms considered here: L1, L2, and $|H|$ -- the hypothesis encoding length term of the MDL objective. $|H|$ is jagged and non-differentiable but results in the correct net being an optimum of the full loss function (Figure~\ref{fig:golden-losses-2d}).}
\label{fig:regularizations-3d}
\end{figure*}

\section{Introduction}
\label{sec:intro}

Probing the capabilities of Artificial Neural Networks (ANNs) in the domain of language learning has advanced in two complementary paths -- theoretical and empirical. Theoretical work tries to delineate the kinds of languages and phenomena that can be expressed by ANNs, and empirical work involves training networks on such tasks and inspecting their performance.

These paths have still not converged: while theoretical work continues to provide findings regarding the expressivity of different architectures, empirical work keeps arriving at suboptimal solutions that fall short of the theoretically correct ones. 
For example, for formal languages such as \ANBN{} or Dyck-1, among many others, we are not aware of any network trained through gradient descent that was shown to perform well on strings that are orders of magnitudes longer than those seen during training, while failures to report generalization beyond low lengths are pervasive (\citealp{JoulinMikolov:2015}, \citealp{WeissGoldbergYahav:2018}, \citealp{SuzgunBelinkovShieberGehrmann:2019}, \citealp{BhattamishraAhujaGoyalWebberCohnHeLiu:2020}, \citealp{El-NaggarMadhyasthaWeyde:2022}, among others; see \citealp{LanChemlaKatzir:2023} for an overview). This stands in contrast to symbolic models, where the requirement for solution correctness across lengths is trivially met.

Why this would be the case is often either left unexplained or waved off as a shortcoming of the optimization method (most often, gradient descent using backpropagation). In this work we argue that these failures are not due to training misfortunes that could be overcome, for example, by using a more exhaustive hyper-parameter search, more training steps, or more training data. Rather, they are due to inherent characteristics of the training objectives currently used for such tasks.

Our main contributions are:
\begin{enumerate}
    
\item We present a manually built, optimal Long Short-Term Memory network (LSTM; \citealp{HochreiterSchmidhuber:1997a}) that accepts the formal language \ANBN{}, following a general recipe given in \citet{WeissGoldbergYahav:2018}. We show that this network would not be found using standard training objectives, since it does not lie at optimum points of these objectives -- even when using regularization terms which according to common wisdom should result in general solutions.

\item We show that by replacing these objectives and regularization terms with an objective to minimize the network's Minimum Description Length (MDL, \citealp{Rissanen:1978}), accompanied by an intuitive encoding scheme, the optimal network becomes an optimum of the objective. 
\end{enumerate}

The full experimental materials and source code are available at \REPOURL{}.

\section{Previous work}
\label{sec:previous-works}

We rely mainly on three recent works, which we extend in the following ways. First, the current work is similar to \citet{El-NaggarRyzhikovDaviaudMadhyasthaWeyde:2023}, who inspected the role of the objective function in formal language learning. They showed that for a simple recurrent neural network (RNN), which uses a single ReLU layer, the optimal counting solution does not align with optima of common loss functions (cross-entropy and mean squared error). This was done by providing necessary and sufficient conditions for implementing counting in a ReLU-RNN. 
We extend this work in the following ways. First, we move to the more commonly used LSTM RNN. Since this architecture is more complex, it is also harder to find such sufficient and necessary conditions for counting as done by \citet{El-NaggarRyzhikovDaviaudMadhyasthaWeyde:2023}. This leaves our results mostly empirical, compared to their analytical result. However, here we go beyond that work by also providing an alternative objective function (MDL), for which the optimal network becomes an optimum.

Second, in order to locate such an optimum of the objective, we build an optimal LSTM that accepts a specific formal language. For this we rely on \citet{WeissGoldbergYahav:2018}, who showed that an LSTM can theoretically implement counting using specific weight configurations, so that the state vector holds a counter that can be incremented and decremented based on the input. Here we implement their general recipe to build an optimum LSTM that accepts the language $a^nb^n$. We focus on one language for simplicity, and the method can be easily extended to more languages.

Third and closest to the current work, \citet{LanGeyerChemlaKatzir:2022} applied the MDL principle to RNNs for formal language learning. The resulting networks were shown to be correct for any string for languages such as \ANBN{}, $a^nb^mc^{n+m}$, and Dyck-1. 
Since this objective resulted in a non-differentiable loss function, \citet{LanGeyerChemlaKatzir:2022} used neuroevolution to search the hypothesis space, evolving free-form RNN cells. Since our focus in this work is the objective, here we leave the search algorithm aside and use a single fixed architecture for which a theoretical target is known to exist (LSTM). We then inspect the effect of the objective function on potential weight solutions.

More broadly, empirical work using RNNs for artificial grammar learning have been carried out at least since the introduction of Simple RNNs in \citet{Elman:1990}. Theoretical work regarding RNNs' theoretical computational power go back at least to \citet{SiegelmannSontag:1992}, who showed that RNNs are Turing-complete under certain permissive assumptions (unbounded activation precision and running time). The empirical success of ANNs in the practical field of natural language processing (NLP) has led to recent interest in the theoretical power of RNNs under practical conditions, mainly real-time processing and finite precision (\citealp{WeissGoldbergYahav:2018}, \citealp{MerrillWeissGoldbergSchwartzSmithYahav:2020}). Other recent work has applied similar methods to the transformer architecture (see survey in \citealp{StroblMerrillWeissChiangAngluin:2023}). 

In terms of empirical results, works since \citet{Elman:1990} trained ANNs to recognize formal languages and most often tested for generalization using unseen string lengths and depths (\citealp{GersSchmidhuber:2001}, \citealp{JoulinMikolov:2015}, among many others).
\citet{LanChemlaKatzir:2023} provide an overview of such works; they show that common to these works is a failure to generalize beyond a certain tested length. \citet{LanChemlaKatzir:2023} also provide a standardized benchmark for formal language learning, and find that RNNs trained to optimize standard losses fail to generalize well from reasonably small amounts of data, while an RNN variant trained to minimize MDL \citep{LanGeyerChemlaKatzir:2022} is able to generalize significantly better.

Applying the MDL criterion to ANNs also dates back to at least the early 1990's. (See \citealp{Schmidhuber:1997} for an overview of early attempts in this area, and \citealp{LanGeyerChemlaKatzir:2022} for a review of more recent work.) \citet{HintonVanCamp:1993} minimized the encoding length of the weights alongside the prediction error, while leaving the architecture fixed. 
\citet{HochreiterSchmidhuber:1994} provided an algorithm that searches for networks that lie at `flat minima' -- regions of parameter space where error remains relatively similar; this preference is given an MDL justification.
\citet{ZhangMuhlenbein:1993} used a genetic algorithm to search for network architectures that minimize an MDL score, using a weight encoding similar to L2 regularization. \citet{Schmidhuber:1997} presented an algorithm for discovering networks that optimize a running-time based complexity metric that is closely related to MDL (Levin complexity). 
More recently, \citet{LouizosUllrichWelling:2017} used a Bayesian method with an MDL justification for pruning and quantizing network units and weights. \citet{LouizosWellingKingma:2018} used a differentiable approximation of the L0 norm to encourage weight sparsity. As far as we can tell, these methods are still prone to the risk of overfitting through highly informative yet small-valued weights, explained in Section~\ref{sec:mdl:encoding} below.

\section{General setup}
\label{sec:general-setup}

We describe here the technical background leading to the experiments in Section~\ref{sec:experiment-golden-anbn}.

\subsection{Minimum Description Length}
\label{sec:mdl}

Striking a balance between model complexity and its fit to the data is important in order to avoid both overfitting and underfitting. It is generally assumed that minimizing model complexity is good (Occam's razor). 

This general principle was formalized within Kolmogorov Complexity (\citealp{Solomonoff:1964}, \citealp{Chaitin:1966}, \citealp{Kolmogorov:1968}), defined as the length of the shortest program that generates specific data. 
Kolmogorov Complexity however is non-computable, a result of the target representation being Turing-complete.
The Minimum Description Length principle (MDL; \citealp{Rissanen:1978}) makes it possible to escape the non-computability by relaxing the requirement of a Turing-complete representation and moving to a less powerful formalism (e.g., a context-free grammar).

Formally, consider a hypothesis space $\mathcal{H}$ and input data $D$. 
The MDL principle aims to find a hypothesis $H^*$ that minimizes the sum:

\begin{equation}
    H^* = \arg\min_{H\in \mathcal{H}} |H|_C + |D:H|
    \label{eq:mdl}
\end{equation}
where $|H|_C$ is the length of $H$ encoded using an encoding scheme $C$ for encoding hypotheses in $\mathcal{H}$. Encoding length is usually measured in bits. 
$|D:H|$ is the encoding length of $D$ given $H$.

Minimizing $|H|_C$ alone would result in a degenerate, over-general model that does not fit the data well. Conversely, minimizing $|D:H|$ alone would result in overfitting. Minimizing both terms together results in a reasonable compromise between generalization and accuracy.

\subsection{Encoding a network}
\label{sec:mdl:encoding}

In this work, hypotheses in $\mathcal{H}$ are taken to be LSTM networks with one linear output layer, followed by a softmax function. 
Here we describe an encoding scheme for such networks which makes it possible to measure their encoding length $|H|$.

We first note that a reasonable encoding scheme for networks should follow an intuitive notion of simplicity in order to penalize overfitting (i.e., lead to larger encoding length). Equating scalar magnitude with simplicity is not enough, since it is still possible to `smuggle' large amounts of information inside very small scalar values. One extreme example is using a fractal encoding in the spirit of \citet{SiegelmannSontag:1992} or \citet{Tabor:2000} which encodes a stack inside a small rational number.\footnote{These works use activation values, not weights, to store such values. However, such a construction still illustrates the difference between information content and scalar magnitude.}\textsuperscript{,}\footnote{Admittedly and as discussed also in \citet{WeissGoldbergYahav:2018}, standard gradient-based methods would most probably not reach such highly specific weight configurations. This does not mean however that such solutions do not exist in the search space and that better search algorithms would not find them.}
However, less sophisticated overfitting is also conceivable using highly specific weights, for example if a model assigns specific probabilities due to sampling artefacts in the training set. 
A reasonable objective should make such memorization worthwhile only if the data justify it, e.g., if it contains many repetitions of the same pattern.

Regularization terms such as L1/L2 are not good enough then, since they would deem, for example, a simple value such as $1$ worse than a smaller yet highly informative value (e.g., $\sum_{1}^{n}\frac{2w_i+1}{4^i}<1$, the fractal encoding of a binary stack $w$, from \citealp{SiegelmannSontag:1992}).

For MDL, the encoding scheme $C$ explained in Section~\ref{sec:mdl} needs to be chosen so that it fulfills the simplicity requirement. We opt for the following encoding scheme, used in \citet{LanGeyerChemlaKatzir:2022}.

A weight $w_{ij}$ is represented as a rational fraction $\frac{n}{m}$. The numerator and denominator are encoded using the prefix-free encoding for integers from \citet{LiVitanyi:2008}:
\[
E(n)=\underbrace{11111\ldots1111}_{\text{Unary enc. of }\lceil{log_2{n}}\rceil\ \ }\underbrace{0}_{\text{Separator}}\underbrace{10101\ldots00110}_{\text{Binary enc. of $n$}}
\]

Both encodings are then concatenated, with an extra bit for the sign. For example, the weight $w_{ij} = +\frac{2}{5}$ would be encoded as:
\[
\underbrace{\underbrace{1}_{+}\underbrace{E(2)=11010}_{2}\underbrace{E(5)=1110101}_{5}}_{w_{ij}}
\]

This encoding fulfills the requirement above: the encoding of very specific or informative weights would be considerably longer than that of intuitively simpler values such as $1$.

In the current setup, the LSTM architectures vary only by the size of the hidden vector and the values of the weights. In order to reliably encode a specific network one needs to encode only the weights of the LSTM cell and output layer, and prepend the size of the hidden vector. The encoding of a specific network would then be:
\[
\underbrace{\underbrace{11011}_{E(\text{hidden size})}\underbrace{\underbrace{11\ldots01}_{\cdots}\underbrace{10\ldots01}_{w_{ij}}\underbrace{11\ldots10}_{\cdots}}_{\text{Weight encoding}}}_{\text{LSTM encoding}}
\]

To calculate $|H|$ for networks trained through backpropagation with floating-point weights, in sections below floats are converted to the closest rational with denominator $m\leq 1000$.\footnote{We use the CPython implementation in \texttt{Fraction.limit\_denominator()} which approximates the closest rational with denominator $\leq1000$.
}

\subsection{Language modeling}
\label{sec:language-modeling}

We use the formal language \ANBN{} as a test case throughout this work, and probe different networks' performance on recognizing it. Strings are drawn from the following probabilistic context-free grammar (PCFG):
\begin{equation}
    \begin{array}{l}
    S \rightarrow \Biggl\{ 
        \begin{array}{lll}
            a Sb &  & 1-p \\
            \varepsilon & & p       
        \end{array}
    \end{array}
\label{eq:pcfg-anbn}
\end{equation}
with $p=0.3$ for all tasks. We use a standard language modeling setup in which the network is fed one symbol at a time, and outputs a probability distribution over the alphabet, predicting the next symbol in the string. Following \citet{GersSchmidhuber:2001}, each string starts and ends with a special symbol.

The training set is sampled by generating strings from (\ref{eq:pcfg-anbn}). The validation set consists of all consecutive strings starting right after the last $n$ in the training set. The validation loss is weighted per-sample so that it follows the same power law distribution induced by (\ref{eq:pcfg-anbn}). The train-validation split in all experiments is 95\%-5\%. In the following sections the training size is 1,000, i.e., a 950-50 split. The maximum $n$ in this training set was 21, so the validation set contained all strings with $22\leq n \leq 71$. The test set in all experiments consisted of all \ANBN{} strings with $1\leq n\leq 1{,}500$.

The network is fed one symbol at a time, and at each step outputs a probability distribution $\hat{p}$ over the alphabet for predicting the next symbol in the string. The baseline loss function we use is the standard cross-entropy loss (CE):

\begin{equation}
    CE(p,\hat{p}) = -\sum_{i=0}^{n}p(c_i)log(\hat{p}(c_i))
\end{equation}
where $n$ is the length of a sequence, $c_i$ is the target symbol at time step $i$, $p(c_i)$ is the target probability at time step $i$ for this symbol, and $\hat{p}(c_i)$ is the probability assigned by the network to this symbol at this time step. In a language modeling setting the target $p(c_i)$ is set to 1, resulting in:

\begin{equation}
    CE(p,\hat{p}) = -\sum_{i=0}^{n}log(\hat{p}(c_i))   
    \label{eq:cross-entropy}
\end{equation}
This sum is then averaged over all time steps for all sequences.

To measure accuracy on the task, we use \textit{deterministic accuracy} (\citealp{JoulinMikolov:2015}, \citealp{LanChemlaKatzir:2023}), defined as the ratio of correct answers at parts of the string that are completely predictable (a correct answer being the network assigning the maximum probability to the correct next symbol). For \ANBN{} strings, this means measuring accuracy at the phase that starts once the first `$b$' appears, including the end-of-sequence symbol. Measuring accuracy at the end-of-sequence symbol turns the task into a strict acceptance task and can distinguish a good network that correctly balances the number of $a$'s and $b$'s, from a degenerate network that, e.g, gets a high deterministic accuracy score simply by only predicting $b$'s.

\subsection{Loss surface exploration}
\label{sec:general-setup:loss-exploration}

Our goal is to test which objectives could lead to optimal solutions. Exhaustive search of the parameter space is infeasible. However, we can explore only parts of the loss space and check if an objective function turns out to favor suboptimal solutions over an optimal one. This would be an indicator that this objective is not suitable for the task (and would lead to reliance on meta-heuristics such as early stopping). We do this by exploring the loss surfaces around an optimal network that solves the task perfectly and around a backpropagation-trained network.

For a given network with parameters $\theta$, and for a loss function $L$, the network's loss is $L(\theta)$ (for some input $x$, in the current case a one-hot vector representing the current symbol, omitted here). For the 2D visualization we use below, the area around a specific network's $\theta$ can then be explored by using two direction vectors $\delta$ and $\eta$, and plotting:

\begin{equation}
f(\alpha,\beta)=L(\theta + \alpha\delta+\beta\eta)
\label{eq:loss-2d-exploration}
\end{equation}

We use the exploration technique by \citet{LiXuTaylorStuderGoldstein:2018}: $\delta$ and $\eta$ are randomized from a Gaussian; then, specific parts of each direction vector are normalized so that they have the norm of the respective parts in the original $\theta$. For fully-connected layers like the ones used in LSTMs, normalization is done for each set of weights leading to a specific neuron. This normalization technique preserves the relative scale of different weight components of a network, and was shown to better reflect properties like convexity when exploring a network's surrounding space. In all plots below we use 51 equally spaced values of $\alpha, \beta \in [-1,1]$. Exploration using larger ranges did not affect the results either visually or quantitatively. 

\subsection{Objectives}

The objective functions for all tasks below share the following structure:

\[
L(\theta) = CE + \lambda Reg(\theta),
\]
Here, $CE$ is the training cross-entropy loss (\ref{eq:cross-entropy}) using the distribution $\hat{p}$ outputted by the network. For the MDL objective in (\ref{eq:mdl}), CE serves as $|D:H|$.
This can be justified in encoding-length terms since (\ref{eq:cross-entropy}) gives the expected length in bits for transmitting the string using Shannon-Fano encoding.

In the second term, $Reg$ is either 
$L_1(\theta) = \sum_{w_{ij}\in \theta} |w_{ij}|$, 
$L_2(\theta) = \sum_{w_{ij}\in \theta} {w_{ij}}^2$, or no regularization. For the MDL objective, $Reg(\theta)$ is $|H|$ -- the encoding length of a network encoded using the method in Section~\ref{sec:mdl:encoding}. $\lambda$ is a coefficient used to calibrate the level of regularization during training, and is usually chosen empirically. 

The common wisdom motivating the regularization term in all cases is to prevent models from overfitting. In the L1/L2 regularization framework, this is done by preventing large weights. Using L1 also leads to a preference for zero-value weights, thus potentially removing connections altogether; this can be thought of as a differentiable way to perform architecture search. However, as suggested in Section~\ref{sec:mdl:encoding}, neither term is a good proxy for the $|H|$ term, since small weights can in fact be very informative (i.e.,~very complex).

Figure~\ref{fig:regularizations-3d} plots the three regularization terms considered here, surrounding the optimal network presented in Section~\ref{sec:experiment-golden-anbn}. It can be seen that while the loss surfaces for L1/L2 are smooth, moving to MDL would result in a highly irregular surface, hostile to gradient methods. 

\section{Optimal vs. trained \ANBN{} LSTM}
\label{sec:experiment-golden-anbn}

\begin{figure}[t!]
    \centering
    \begin{subfigure}{.47\textwidth}
        \includegraphics[width=\textwidth]{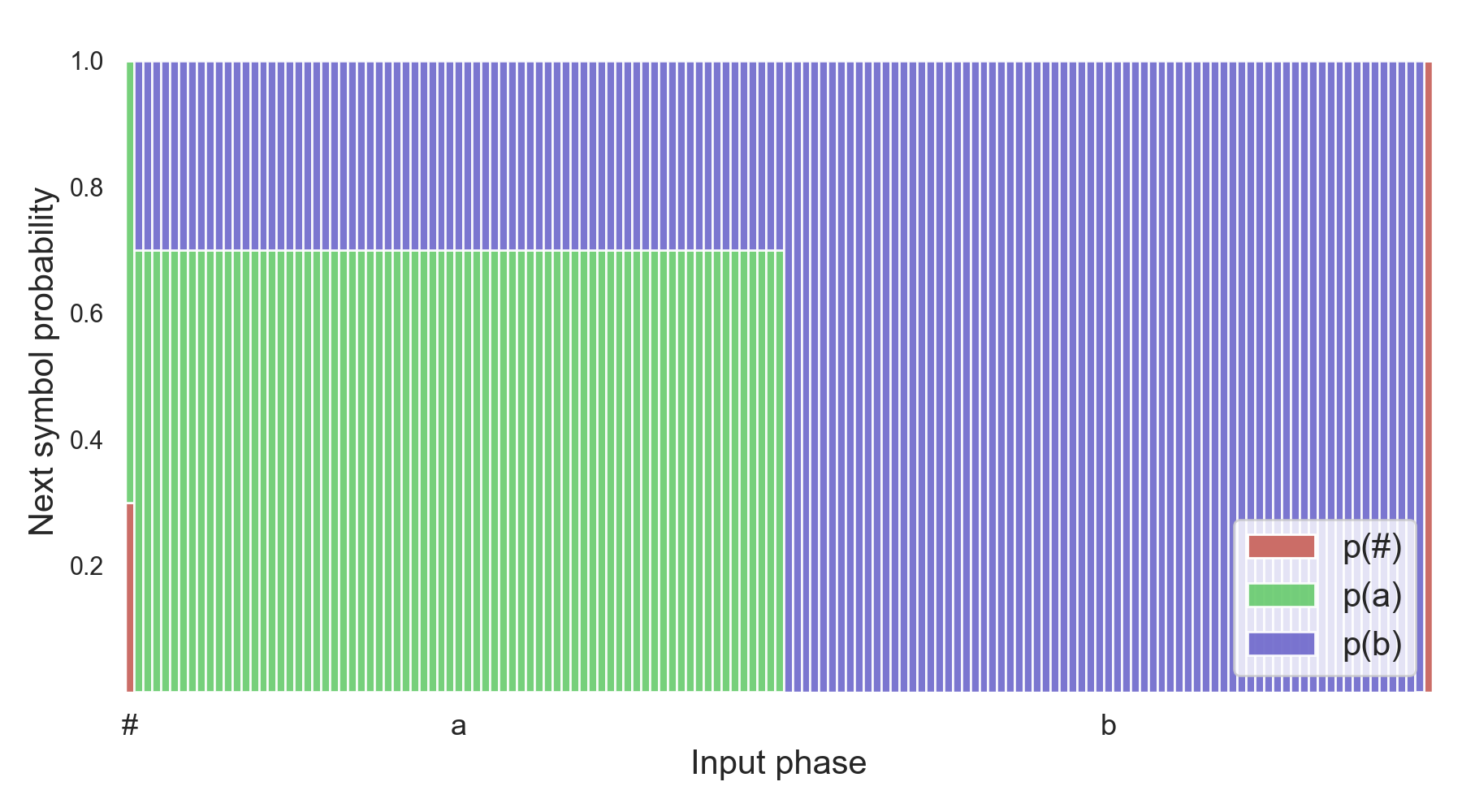}
        \caption{Golden network}\label{fig:output:probabs:golden}
    \end{subfigure} \hfill
    \begin{subfigure}{.47\textwidth}
        \includegraphics[width=\textwidth]{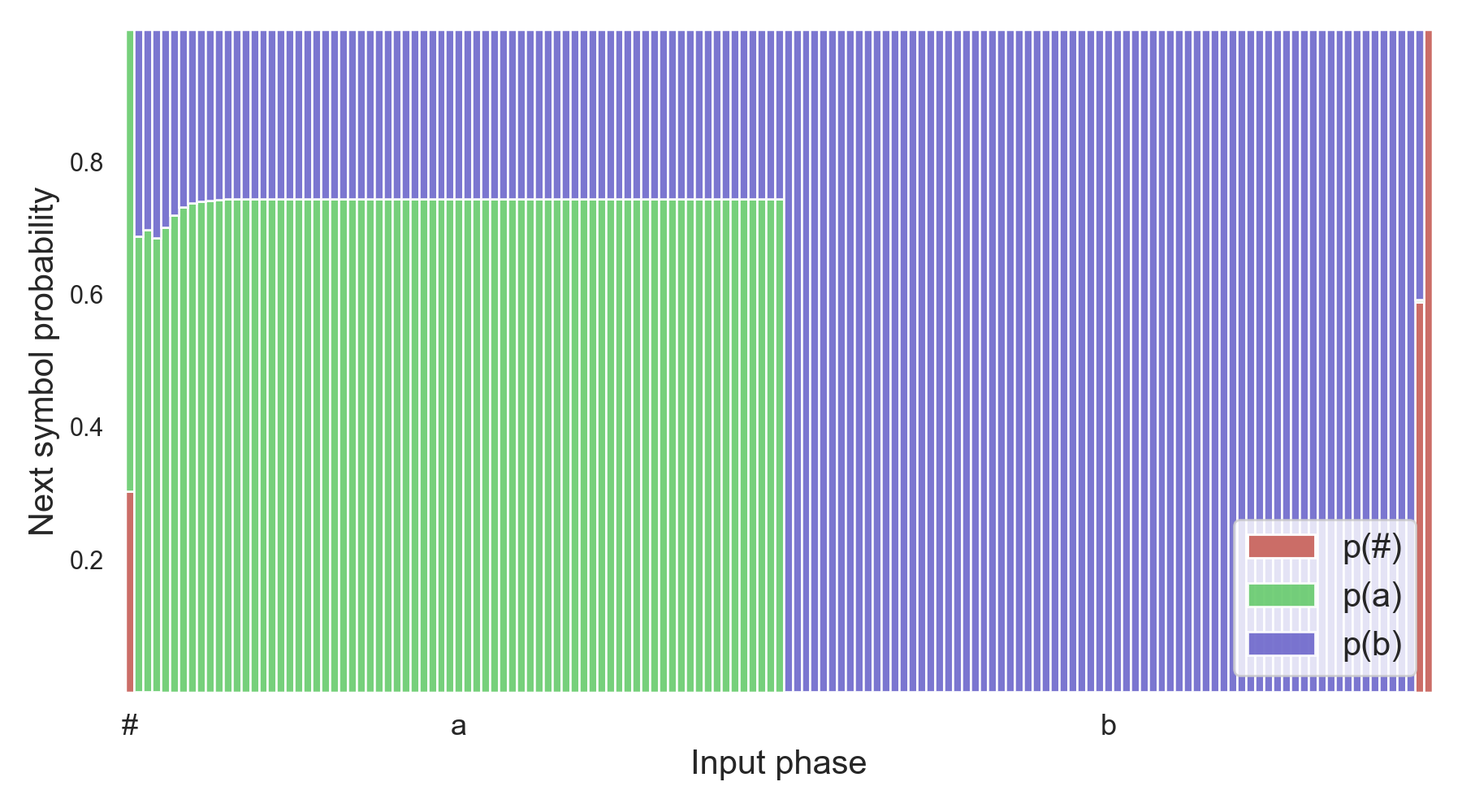}
    \caption{Best trained network}
    \label{fig:output:probabs:trained}
    \end{subfigure}
    \caption{Output probabilities assigned by the golden network and the best trained network for $n=73$, the first point of failure of the trained network. Going left to right, each column represents the probability distribution outputted by the network at each time step.}
    \label{fig:output:probabs}
\end{figure}

Here we compare an optimal, manually-constructed LSTM that recognizes the language \ANBN{} perfectly, with an LSTM trained through backpropagation on the same task. We name the optimal network `golden' to avoid confusion with general optimum points.
The golden network is optimal in the sense that it always outputs the correct probabilities at each step of an \ANBN{} string drawn from (\ref{eq:pcfg-anbn}), for any value of $n$. The optimal probabilities are presented in Figure~\ref{fig:output:probabs:golden}.

The goal of the experiment is to test whether a perfect solution can be found when using the different objectives considered here. Exhausting the entire parameter space is infeasible, even for networks with the very small hidden size (3) used here. However, if the golden network turns out to not be an optimum of certain objectives, i.e., a worse-performing network will be deemed better by an objective, we can conclude that this objective would not lead to this specific network.

\subsection{Golden \ANBN{} network}

The golden network is implemented based on a general recipe given in \citet{WeissGoldbergYahav:2018}, who showed that an LSTM can theoretically implement counting using specific configurations of the gate weights, so that the state vector holds a counter that can be incremented and decremented based on the input. This makes it possible for LSTMs to recognize a family of formal languages called Counter Languages. Roughly, this family corresponds to languages which can be recognized using a counting mechanism in real-time \citep{Merrill:2021}. This includes \ANBN{}. In an empirical experiment, however, \citet{WeissGoldbergYahav:2018} trained LSTMs on recognizing the language and found that the networks did not converge on the fully general counting solution, rather it converged on a suboptimal solution that failed to recognize \ANBN{} strings starting at $n$ as low as 256. 

\begin{figure}[t!]
    \centering
    \begin{subfigure}{.47\textwidth}
        \includegraphics[width=\textwidth]{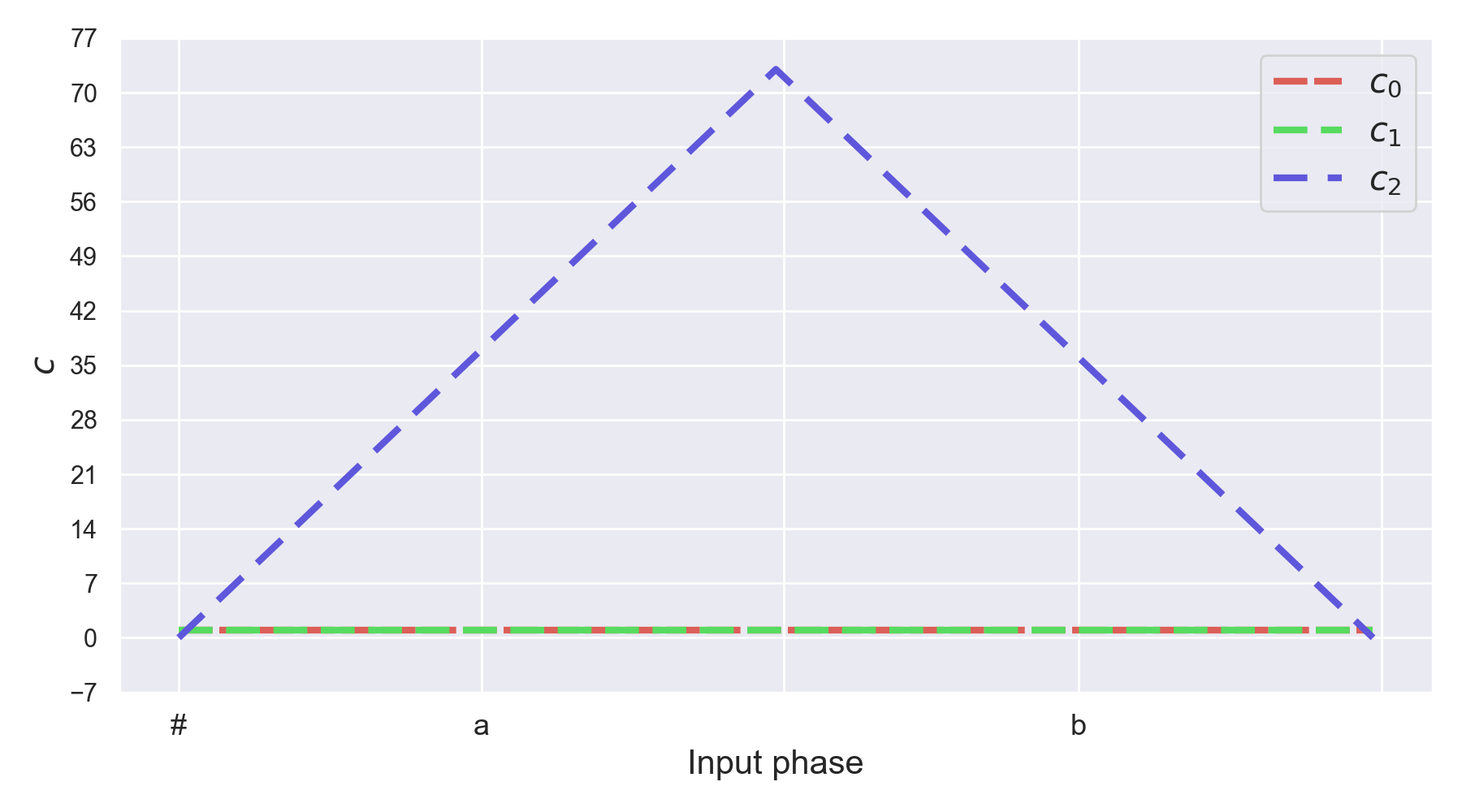}
        \caption{Golden network}\label{fig:output:memory:golden}
    \end{subfigure} \hfill
    \begin{subfigure}{.47\textwidth}
        \includegraphics[width=\textwidth]{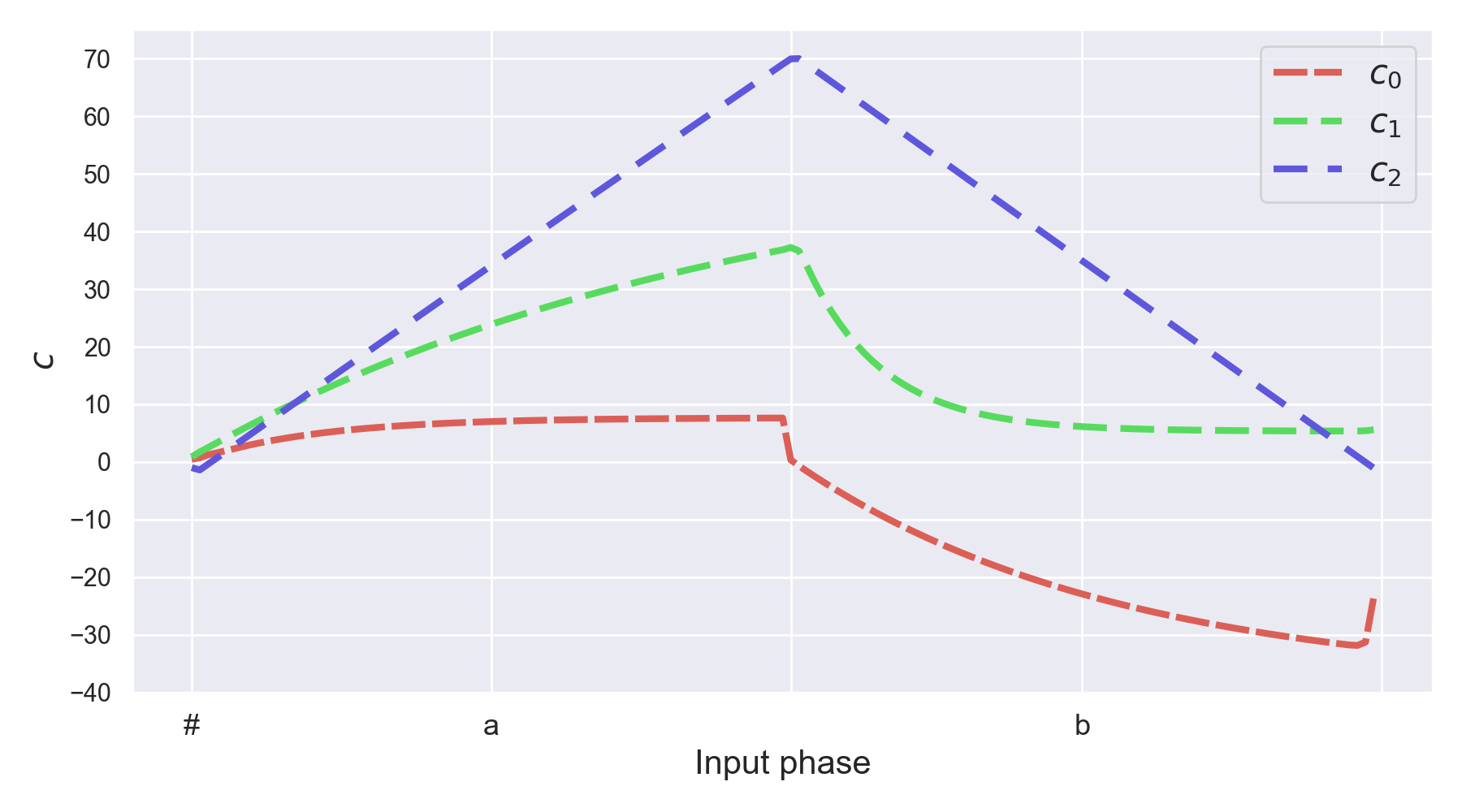}
    \caption{Best trained network}
    \label{fig:output:memory:trained}
    \end{subfigure}
    \caption{Memory values for the golden network and the best trained network for $n=73$, the first point of failure of the trained network. Each line corresponds to one component of $c$, the memory vector in the LSTM cell, as the string is fed to the network.
    }
    \label{fig:output:memory}
\end{figure}

We describe here in general terms the mechanics of the golden network. The full construction is given in Appendix~\ref{appendix:anbn-net-mechanics}. The weights of the LSTM cell are set so that the network keeps track of the number of $a$'s compared to the number of $b$'s seen at each time step. Figure~\ref{fig:output:memory:golden} plots the values of the memory vector of the LSTM during feeding of an \ANBN{} string, illustrating its counting mechanism. On top of the LSTM cell we add a linear layer that receives the hidden state as input, and outputs the correct target probabilities through a final softmax.

The manual network reaches 100\% test accuracy on the test set which contains all $a^nb^n$ strings with $1\leq n\leq 1500$, and in fact can be shown to be correct for any $n$. Note that the network is optimal with respect to its performance, and that other perfect networks with better MDL scores could exist in the search space. We return to this point in the Limitations section.

\subsection{Backpropagation-trained \ANBN{}\ LSTM}
\label{sec:experiment-backprop}

We compare the golden network with networks trained on the same task using standard techniques. We run a hyper-parameter grid search to train LSTMs that have the same architecture as the manual network: hidden size 3 and a single linear output layer, followed by a softmax. The grid covers the following hyper-parameters: training set size, weight initialization method, regularization term, dropout rate, and early stopping patience based on validation loss, across five different random seeds. The grid yields 3,360 configurations. The full hyper-parameter grid is given in Appendix~\ref{appendix:grid-hyper-params}.

The overfitting prevention techniques explored here belong to two groups: techniques where a regularization term is added directly to the loss function (L1, L2), and meta-heuristics external to the objective (dropout, early-stopping based on validation loss), aimed at preventing the loss from getting too low. While our focus here is the objective function, 
we still include dropout and early stopping in order to compare the golden network with a network trained in a practical setting with the best chances to succeed. In Table~\ref{table:exhaustive-lambdas} in the appendix we provide the same comparison for the best network that was trained without early stopping or dropout.

We select the best network out of all runs based on validation loss. The best network reached 77.3\% deterministic test accuracy (on $1\leq n\leq 1{,}500$) and was trained with the following parameters:
training size 1000 (950-50 train-validation split); no regularization term; early stopping patience of 2 epochs; no dropout; normal weight initialization.

\subsection{Network behavior}

We start by comparing the behavior of the golden and trained networks. The best trained network correctly accepts all strings with $n\leq72$ (the largest $n$ in the training set was 21). We compare the networks using the first point of failure of the trained network, $n=73$.
Figure~\ref{fig:output:probabs} plots the output probabilities assigned by the two networks throughout the sequence, and Figure~\ref{fig:output:memory} plots their memory values ($c$ in the LSTM cell).

We first examine the network outputs in Figure~\ref{fig:output:probabs}.
At first blush, the trained network seems successful, following the language distribution induced by (\ref{eq:pcfg-anbn}) and visualized in Figure~\ref{fig:output:probabs:golden} almost perfectly. 
However, the network is imperfect in two ways: first, probabilities at the beginning of the string are incorrect, most probably due to overfitting of more frequent low-$n$ values in the training set. Additionally and more crucially, the network's count seems to leak, with probability mass for the end-of-sequence symbol assigned to the before-last time step.
This becomes a problem for $n\geq73$, when the network starts accepting illicit $a^nb^{n-1}$ strings.

As for the inner workings of the network, visualizing the network's memory in Figure~\ref{fig:output:memory:trained} shows that the network has indeed developed some counting mechanism in at least one component of the memory vector ($c_0$), which seems to be imperfect as it does not correctly count up to 73, and goes slightly below 0 towards the end of the string. (It is unclear however how the network uses the other two components, which could potentially complement this counter).

\subsection{Loss exploration}
\label{sec:loss-exploration}

\begin{figure*}[t]
    \centering

    \begin{subfigure}{.32\textwidth}
    \centering
    \includegraphics[width=\textwidth]{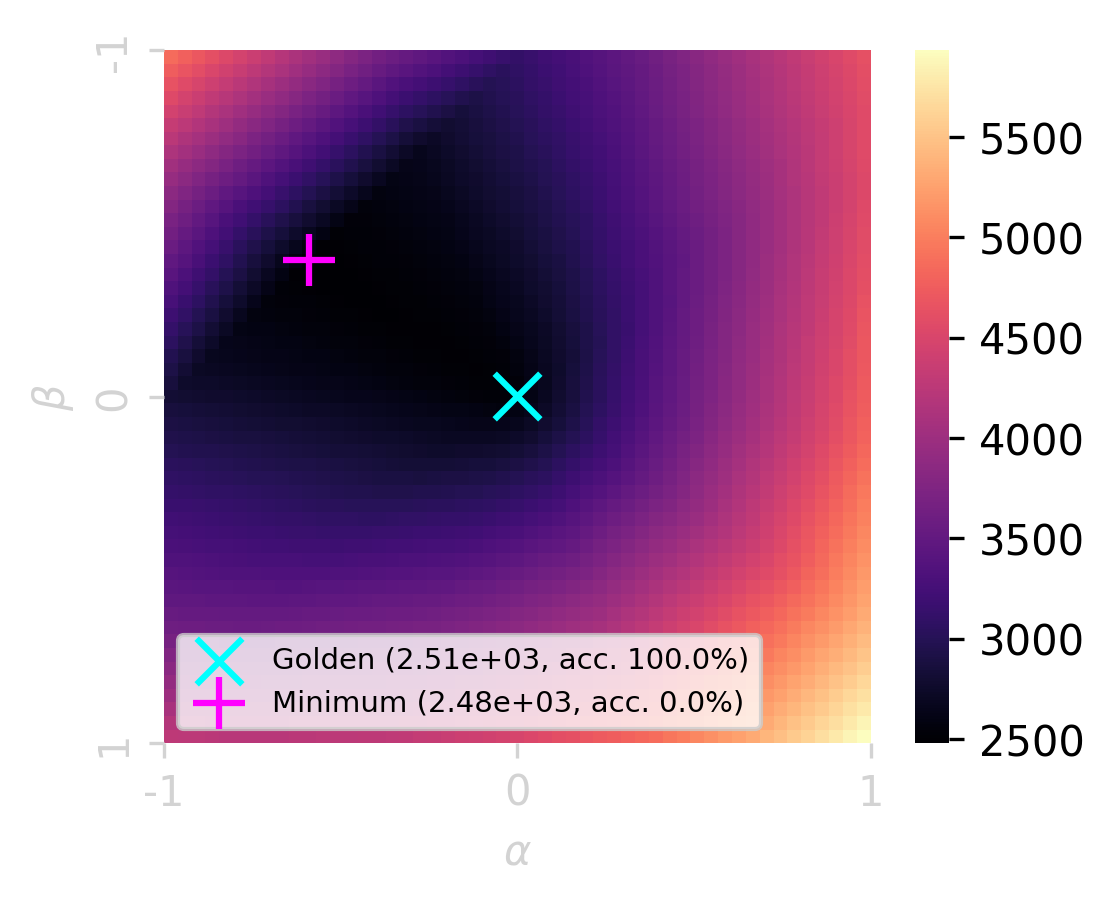}
    \caption{$CE_{train} + L1$}
    \label{fig:golden-losses-2d-l1}
    \end{subfigure}\hfill
    \begin{subfigure}{.31\textwidth}
    \centering
    \includegraphics[width=\textwidth]{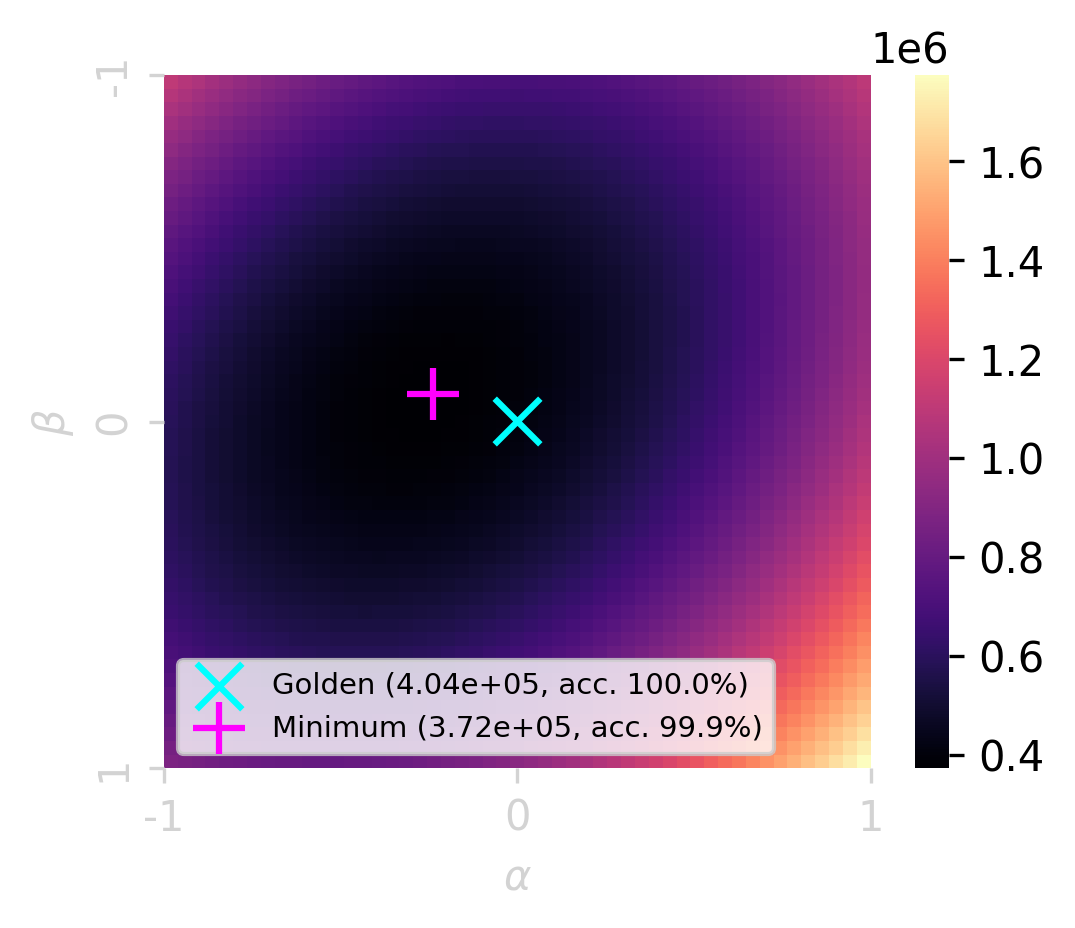}
    \caption{$CE_{train} + L2$}
    \label{fig:golden-losses-2d-l2}
    \end{subfigure}\hfill
    \begin{subfigure}{.34\textwidth}
    \centering
    \includegraphics[width=\textwidth]{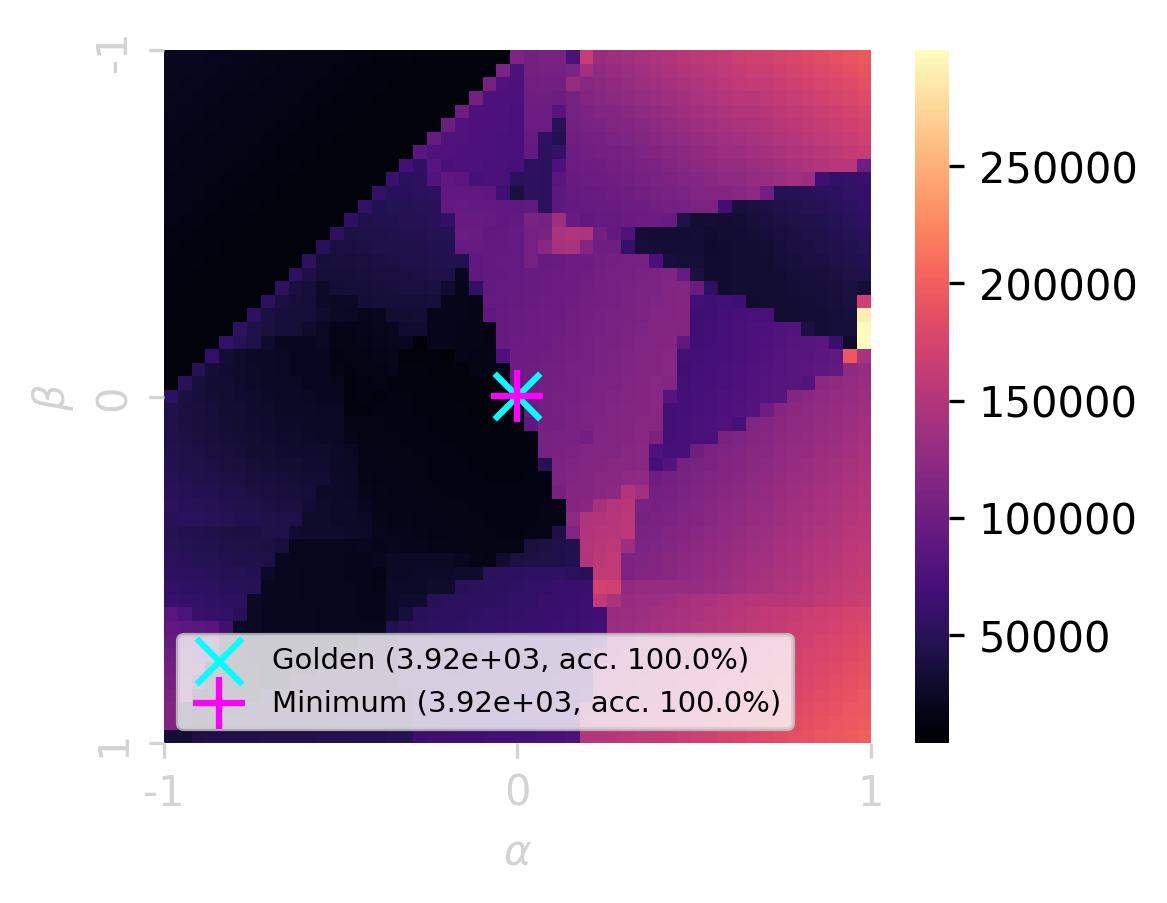}
    \caption{$CE_{train} + |H|\ \text{(MDL)}$}
    \label{fig:golden-losses-2d-mdl}
    \end{subfigure}\hfill

\caption{Training loss around the golden \ANBN{} LSTM, and test accuracy scores for the golden network and the local minimum network.
Optimizing using L1 or L2 (\ref{fig:golden-losses-2d-l1}, \ref{fig:golden-losses-2d-l2}) would result in suboptimal networks, while MDL (\ref{fig:golden-losses-2d-mdl}) results in alignment of the golden network with an optimum point of the loss.
}
\label{fig:golden-losses-2d}
\end{figure*}

\begin{figure*}[t]
    \centering

    \begin{subfigure}{.33\textwidth}
    \centering
    \includegraphics[width=\textwidth]{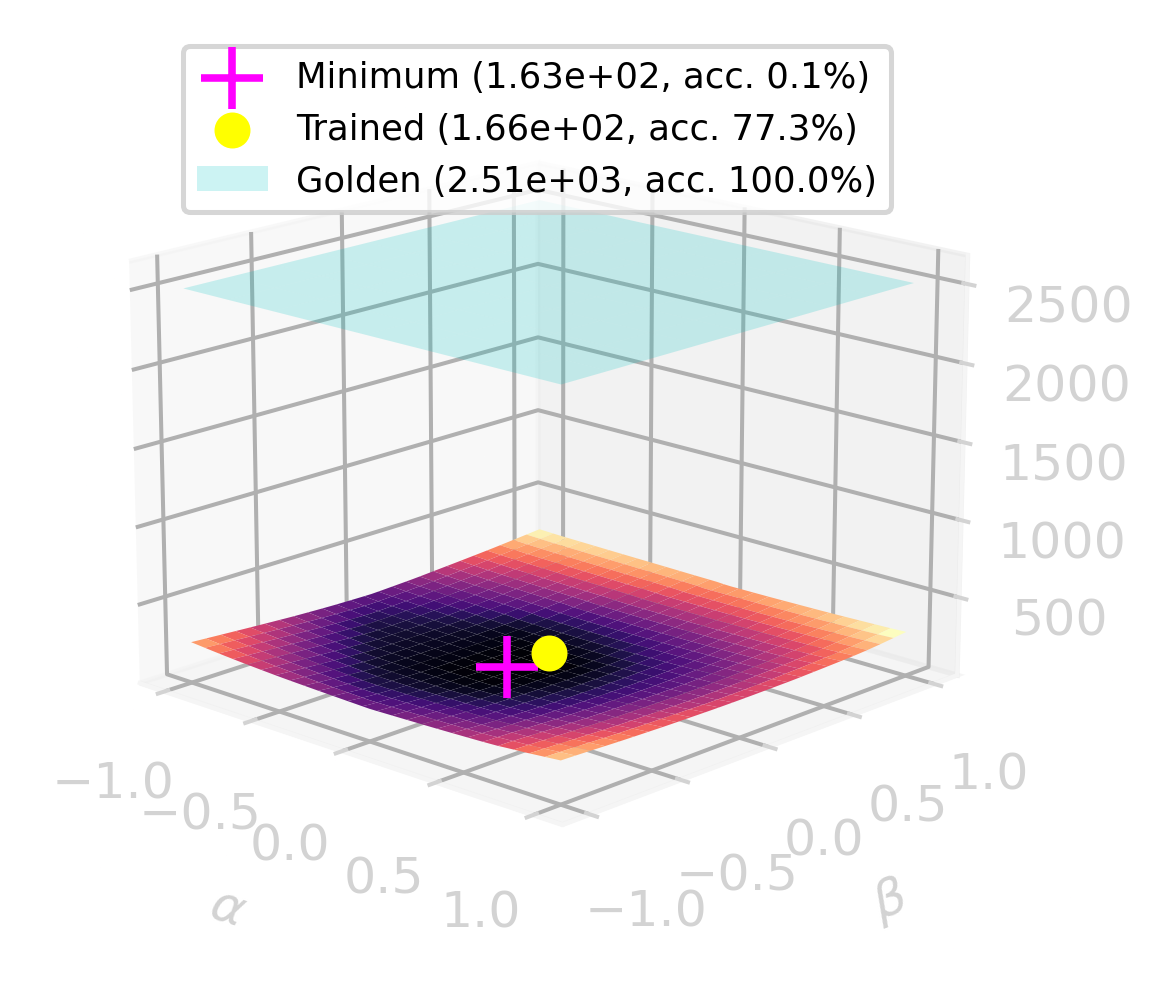}
    \caption{$CE_{train} + L1$}
    \label{fig:best-trained-losses-3d-l1}
    \end{subfigure}\hfill
    \begin{subfigure}{.33\textwidth}
    \centering
    \includegraphics[width=\textwidth]{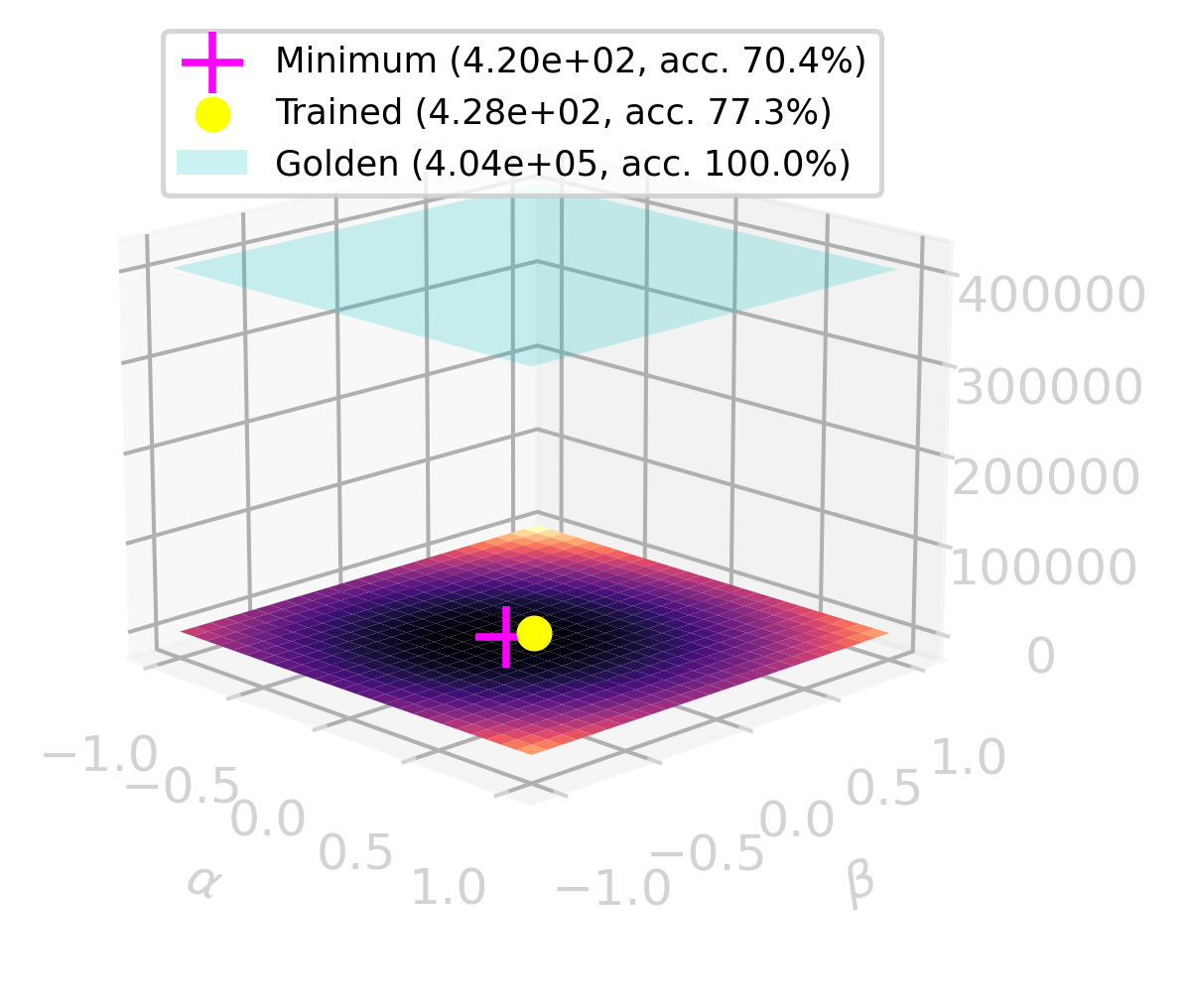}
    \caption{$CE_{train} + L2$}
    \label{fig:best-trained-losses-3d-l2}
    \end{subfigure}
    \begin{subfigure}{.33\textwidth}
    \centering
    \includegraphics[width=\textwidth]{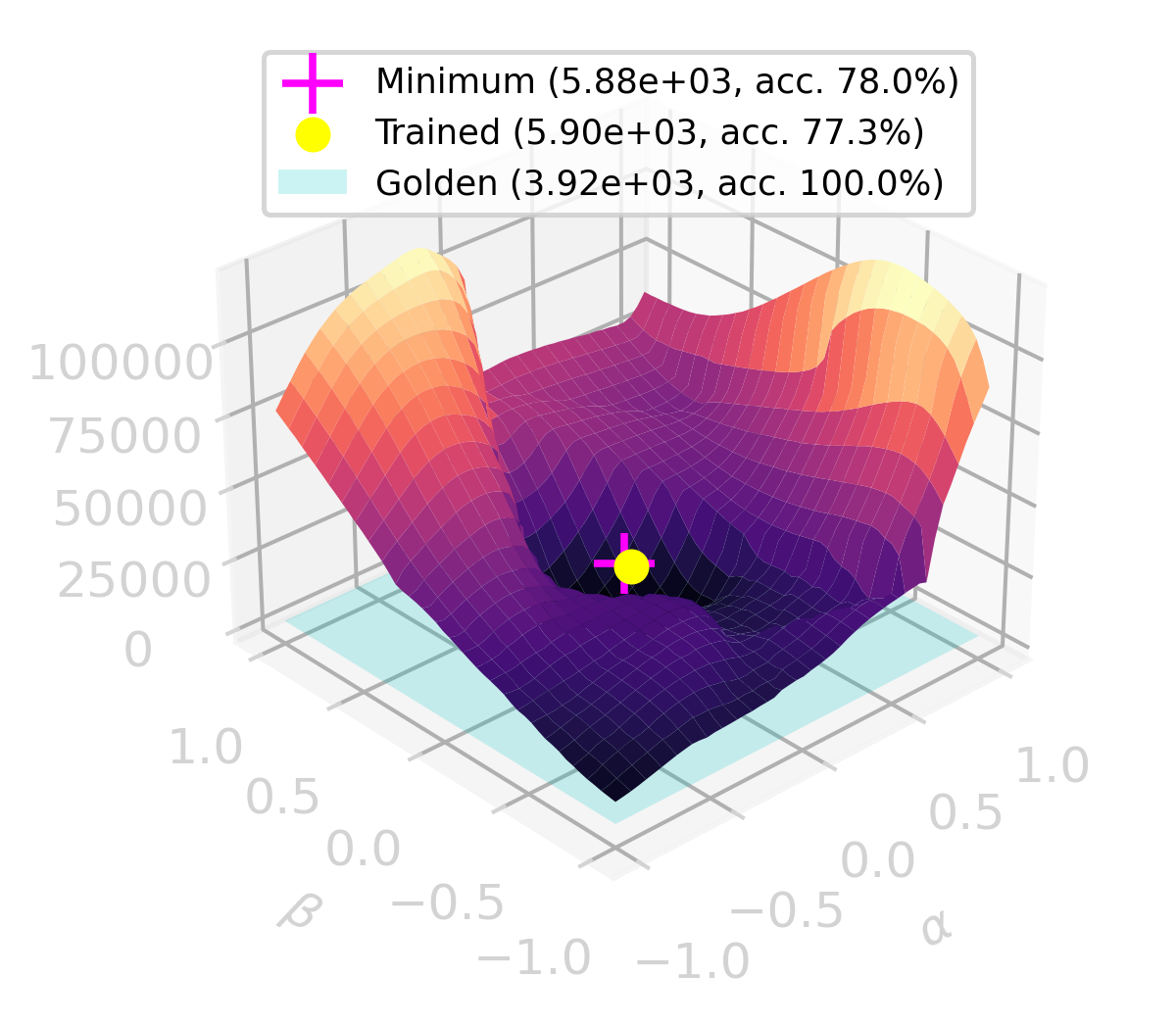}
    \caption{$CE_{train} + |H|\ \text{(MDL)}$}
    \label{fig:best-trained-losses-3d-mdl}
    \end{subfigure}\hfill

\caption{Training loss surfaces and the test accuracy of the best \ANBN{} LSTM found through a hyper-param grid search, trained using backpropagation with the standard cross-enropy loss and early stopping based on validation loss. When evaluated using L1 or L2 (\ref{fig:best-trained-losses-3d-l1}, \ref{fig:best-trained-losses-3d-l2}) the network ranks better than the golden network but has worse test performance. When evaluated using MDL (\ref{fig:best-trained-losses-3d-mdl}) the trained network does not minimize the loss as well as the golden network, ending up in a smooth but suboptimal area of the loss space.}
\label{fig:best-trained-losses-3d}
\end{figure*}

Is the suboptimal performance of the trained network above simply a misfortune of the current setup? We explore the possibility that the culprit might be the objective function. We do this by comparing the loss values of the golden network with the trained network's, using standard objectives and the MDL objective. 

Beyond measuring the loss value of the two networks considered here, we also explore their surrounding loss landscape in order to check for alternative local minima and inspect properties like convexity and smoothness of the loss. This is done using the technique described in Section~\ref{sec:general-setup:loss-exploration}.

Figures~\ref{fig:golden-losses-2d} and~\ref{fig:best-trained-losses-3d} plot the different loss surfaces around the networks. On each plot we mark the minimum point in the neighborhood, to check if it aligns with the network under investigation. If it does not, using that loss (either for fine-tuning the network or training from scratch) would potentially end up at that other minimum. For each relevant point we use the parameter vector to build the corresponding LSTM and calculate its test accuracy.

We start by exploring the loss surface around the golden network.\footnote{We omit plotting the standalone cross-entropy loss because it is trivial to show that minimizing this loss alone will lead to overfitting (partially explaining the fact that the best performing network ends up using early stopping).
Table~\ref{table:exhaustive-lambdas} in the appendix demonstrates this using the next-best grid-search network that was trained without a regularization term or early stopping, whose cross-entropy loss goes below that of the golden network's.} Figures~\ref{fig:golden-losses-2d-l1} and \ref{fig:golden-losses-2d-l2} show that if L1 or L2 regularization were used, the golden network would not have been found -- rather, using these regularization terms would lead the search to suboptimal networks that have better training loss values, but also worse test accuracy.

For the MDL objective function, visualized in Figure~\ref{fig:golden-losses-2d-mdl}, the minimum in the plotted area aligns with the golden network, showing that at least in this neighborhood, searching with MDL as an objective would lead to the correct solution. In Section~\ref{sec:limitations} below we discuss potential limitations to these findings.

Figure~\ref{fig:best-trained-losses-3d} plots the different loss surfaces around the best trained network. We plot in 3D for comparison with the relevant value for the golden network, which lies in a different area of the loss space. 
Here, for all objectives, the winning network lies in a smooth and convex area. 
When evaluated using L1 and L2 regularization, the golden network ranks worse by the relevant losses. 

For the MDL objective the image is reversed: the trained network ranks worse, while the MDL score of the golden network remains unreachable below (Figure~\ref{fig:best-trained-losses-3d-mdl}). Since the two networks' cross-entropy terms are almost identical (see Table~\ref{table:exhaustive-lambdas}), this inversion is mainly due to to the $|H|$ term, which suggests that the trained network uses over-informative weights.
Results for more $\lambda$ values for all networks are given in Table~\ref{table:exhaustive-lambdas}. 

\section{Discussion}
\label{sec:discussion}

We presented a comparison between common overfitting-prevention techniques, among them some that equate simplicity with scalar magnitude, and the MDL objective accompanied with an encoding scheme for weights which favors an intuitive notion of simplicity. 
Combined with a  manually built LSTM that optimally recognizes \ANBN{}, we could measure the loss values of an optimal solution and check if they align with optimum points of the loss function. It was only when we used MDL that the optimal network aligned with the minimum of the loss. For the other loss functions, networks lying at minimum points had far from optimal performance. Using meta-heuristics such as early stopping mitigated overfitting to some extent, but still did not lead to a fully general solution.

We interpret these findings as an indicator that ANNs' failure to converge on provably existing, optimal solutions is not accidental, but rather an inherent and pathological property of the way that current models are trained. This is in line with a mounting list of generalization failures to learn formal languages, as well as more complicated natural language tasks.

We focused here on RNNs, mainly because they lend themselves easily to manual construction and inspection. However, we see no \emph{a priori} reason why our results would not extend to other architectures such as transformers or convolutional networks, given the generality of the MDL principle, and the fact that it has been shown to be beneficial across various domains and learning tasks, including linguistic phenomena (see \citealp{Stolcke:1994}, \citealp{Grunwald:1996}, \citealp{Marcken:1996}, and \citealp{RasinBergerLanShefiKatzir:2021}, among others).

\section{Limitations}
\label{sec:limitations}

In Section~\ref{sec:loss-exploration} we explored the loss surface around the golden network using different objectives, and found that using L1/L2 regularization leads to suboptimal networks lying at optimum points, while using the MDL objective leads to the golden network lying at an optimum point. However, since the loss exploration is not (and cannot be) exhaustive, caution is needed when making generalizations based on these results.

First, when using L1/L2, it is still possible of course that better optima lie somewhere else in the loss spaces, and that the respective minimizing networks have perfect performance. Given the discussion in Section~\ref{sec:mdl} about scalar magnitude vs. simplicity we find this possibility unlikely but admittedly still possible.

Conversely, when using the MDL objective, here it is conceivable that other networks would have better MDL scores and suboptimal performance. While this cannot be ruled out completely, we believe that using the MDL objective accompanied by a reasonable encoding scheme like the one used here makes over/under-fitting unlikely. This is arguably not the case for L1/L2. (Another possibility, that of a network with a better MDL score but equivalent perfect performance, is more likely, given that the golden network was manually designed and can potentially be optimized further.)

Finally, a major practical limitation of the current work relates to the non-differentiability of the MDL objective. This is especially problematic for ANNs, given that current standard training methods rely almost exclusively on gradient descent. One could then consider L1/L2 as a differentiable proxy for a strict formalization of simplicity. However, the current work sheds light on the shortcomings of these compromises. This in turn could lead both to a more informed use of such proxies, and potentially to further research regarding better optimization techniques for MDL.

\section{Acknowledgements}

This research was supported by ANR-17-EURE-0017 and ISF 1083/23. This project was provided with computer and storage resources by GENCI at IDRIS thanks to the grant 2023-AD011013783R1 on the supercomputer Jean Zay's V100 partition.

\bibliography{MDL-Regularization}
\bibliographystyle{acl_natbib}

\appendix

\begin{table*}[t]
\centering
\begin{tabular}{llcccccc}
\toprule
\multirow{2}{*}{Loss} & \multirow{2}{*}{$\lambda$} & \multicolumn{2}{c}{Golden net} & \multicolumn{2}{c}{Best trained net} & \multicolumn{2}{c}{Best trained, no early stopping} \\
 & & Loss & Test acc. \% & Loss & Test acc. \% & Loss & Test acc. \% \\
\midrule
CE & - & 3.58e-01 & \ColorBestAcc\textbf{100.00} & 3.58e-01 & 77.33 & \ColorBestLoss\textbf{3.57e-01} & 64.97 \\
\midrule
\multirow{3}{*}{CE + L1} & 0.1 & 2.50e+02 & \ColorBestAcc\textbf{100.00} & \ColorBestLoss\textbf{1.67e+01} & 77.21 & 1.85e+01 & 86.73 \\

 & 0.5 & 1.24e+03 & 96.23 & \ColorBestLoss\textbf{8.17e+01} & 77.21 & 8.94e+01 & \ColorBestAcc\textbf{96.24} \\
 & 1.0 & 2.48e+03 & 0.00 & \ColorBestLoss\textbf{1.63e+02} & 0.13 & 1.78e+02 & \ColorBestAcc\textbf{96.24} \\
 
\midrule
 
\multirow{3}{*}{CE + L2}  & 0.1 & 3.72e+04 & \ColorBestAcc\textbf{99.87} & \ColorBestLoss\textbf{4.23e+01} & 77.21 & 5.80e+01 & 90.34 \\
 & 0.5 & 1.86e+05 & \ColorBestAcc\textbf{99.87} & \ColorBestLoss\textbf{2.10e+02} & 70.38 & 2.85e+02 & 91.81 \\
 & 1.0 & 3.72e+05 & \ColorBestAcc\textbf{99.87} & \ColorBestLoss\textbf{4.20e+02} & 70.38 & 5.69e+02 & 91.81 \\
 \midrule
  MDL & - & \ColorBestLoss\textbf{3.92e+03} & \ColorBestAcc\textbf{100.00} & 5.88e+03 & 77.98 & 5.87e+03 & 69.68 \\
\bottomrule
\end{tabular}
\caption{Minimum training loss values and best test deterministic accuracy scores in the space surrounding the following networks: the golden network, the best trained network which used early stopping, and the best trained network that was trained without early stopping or regularization terms. Winning values across networks are indicated for each row: best loss (blue) and best accuracy (red). Minimizing the loss and achieving perfect accuracy coincide only for the MDL objective.}
\label{table:exhaustive-lambdas}
\end{table*}

\section{Grid search hyper-params}
\label{appendix:grid-hyper-params}

Training size: 500/1000/5000/10000. Random seed: 100/200/300/400/500. Regularization: none/L1/L2. Regularization lambda when relevant: 0.1/0.5/1.0. Dropout rate: none/0.2/0.4/0.6. Early stop after no improvement for number of epochs: none/2/10. Weight initialization: uniform/normal. 

All simulations used the Adam optimizer \cite{KingmaBa:2017} with learning rate 0.001, $\beta_1 = 0.9, \beta_2 = 0.999$, and ran for 20,000 epochs unless stopped by early stopping.

\section{Golden \ANBN{} LSTM construction}
\label{appendix:anbn-net-mechanics}

This section spells out the construction of the optimal \ANBN{} network from Section~\ref{sec:experiment-golden-anbn}.
The network is designed to output the correct probability distribution for \ANBN{} strings induced by the PCFG in (\ref{eq:pcfg-anbn}). The target probabilities are plotted in Figure~\ref{fig:output:probabs:golden}. 

The general idea is to implement a counting mechanism in the LSTM cell and then to pass this value through a linear layer and a softmax, which outputs the target probabilities. 
A full PyTorch implementation of the network is given at \REPOURL{}.

\subsection{Representations and constants}

We use a standard LSTM cell represented by the following functions:

\begin{align}
    & i_t = \sigma(W_{ii} x_t + b_{ii} + W_{hi} h_{t-1} + b_{hi}) \\
    & f_t = \sigma(W_{if} x_t + b_{if} + W_{hf} h_{t-1} + b_{hf}) \\
    & \label{eq:gt:def} g_t = \tanh(W_{ig} x_t + b_{ig} + W_{hg} h_{t-1} + b_{hg}) \\
    & \label{eq:ot:def} o_t = \sigma(W_{io} x_t + b_{io} + W_{ho} h_{t-1} + b_{ho}) \\
    & \label{eq:ct:def} c_t = f_t \odot c_{t-1} + i_t \odot g_t \\
    & h_t = o_t \odot \tanh(c_t)
\end{align}
where $\sigma$ is the sigmoid activation and $\odot$ is the element-wise product.

In the following construction, all weights are set to 0 unless mentioned otherwise.

The LSTM gates (sigmoids and tanh's) need to be saturated in order to prevent leakage and keep the solution stable \cite{WeissGoldbergYahav:2018}. For this, a large enough input needs to be used. We select empirically: 
\[LARGE = 2^7 - 1\]
which is the largest unsigned integer that fits in 7 bits (instead of, e.g., $2^7$, in order to save bits when using the encoding scheme from Section~\ref{sec:mdl:encoding}).

Network inputs and outputs are vectors of size 3, with the following class positions: $[\#, a, b]$, where \# is the start/end-of-sequence symbol. Inputs are one-hot encoded, so that: 
\begin{equation}
x_t = [\mathbbm{1}_\#, \mathbbm{1}_a, \mathbbm{1}_b]
\label{eq:xt}
\end{equation}

The notation $[ \cdots ]_{tanh}$ is used as a shorthand for $tanh([ \cdots ])$. Column vectors are printed as row vectors and omitting the transpose for readability.

\subsection{Counting}

The network's memory vector $c_t$ is of size 3. We describe the construction that leads to $c_t$ holding the following target values at each time step:
\begin{equation}
c_t = [1, 1, \#a-\#b]    
\label{eq:ct}
\end{equation}
where $\#a$ and $\#b$ represent the number of $a$'s and $b$'s seen so far. $\#a-\#b$ thus counts the number of unmatched $a$'s and becomes 0 only when the $a$'s and $b$'s are balanced. The two constant $1$'s will be used downstream.

We first set:

\[
W_{ig} = LARGE \cdot\begin{pmatrix}
1 & 0 & 0 \\
1 & 0 & 0 \\
0 & 1 & -1 
\end{pmatrix} 
\]

Then, from definitions (\ref{eq:gt:def}) and (\ref{eq:xt}) and because all other weights feeding $g_t$ are 0, we get:

\begin{equation}
g_t = tanh(W_{ig}x_t) = [\mathbbm{1}_{\#}, \mathbbm{1}_{\#}, \mathbbm{1}_a - \mathbbm{1}_b]
\label{eq:gt}
\end{equation}
i.e., the last component of $g_t$ holds $+1$ when seeing `$a$` or $-1$ when seeing `$b$`. $1$'s are stored in the other components when the start-of-sequence symbol first appears.

Then, the input and forget gates are saturated to make the addition between $g_t$ and $c_{t-1}$ stable.
Saturating the gates is done through their biases in order to save on encoding length:
\[
b_{ii} = b_{if} = LARGE \cdot [1,1,1]
\]

We write off the saturated gates from definition (\ref{eq:ct:def}), and get the recurrent update of the memory vector:

\begin{equation}
c_t = c_{t-1} + g_t
\label{eq:ct:addition}
\end{equation}

It is then simple to apply the recurrence and get the correct counting targets (\ref{eq:ct}) for all time steps: 
(\ref{eq:gt}) gives $g_0 = [1,1,0]$ for the first time step and $g_{t>0} = [0,0,\mathbbm{1}_a - \mathbbm{1}_b$] for all other steps. 

\subsection{Hidden vector}

The following construction leads to the hidden vector $h_t$ holding the following target values, representing the different phases of an \ANBN{} string:

\begin{equation}
h_t = [\mathbbm{1}_\#, \mathbbm{1}_a, \mathbbm{1}_{\#a>\#b}]_{tanh}
\label{eq:ht:target}
\end{equation}

We first construct $o_t$ as a mask vector to select the relevant part from $c_t$ in (\ref{eq:ct}) based on the current phase of the string. 

The mask $o_t$ is constructed by setting:

\[
W_{io} = LARGE \cdot \begin{pmatrix}
    2 & 0 & 0 \\
    0 & 2 & 0 \\
    0 & 0 & 2 \\ 
\end{pmatrix}
 \]
\[
b_{io} = LARGE \cdot  [-1, -1, -1]
\]
Since all other weights in definition (\ref{eq:ot:def}) are 0, we get:
\[
o_t = \sigma(W_{io}x_t + b_{io})
\]

Following the indexing of $x_t$, this results in a one-hot mask based on the current input: 
\[
o_t = [\mathbbm{1}_\#, \mathbbm{1}_a, \mathbbm{1}_{b}]
\]
($W_{io}$ and $b_{io}$ are used instead of setting $W_{io}$ to the identity because of the sigmoid in (\ref{eq:ot:def}).)

Combined with (\ref{eq:ct}) we get: 
\begin{equation}
h_t = o_t \odot tanh(c_t)  = [\mathbbm{1}_\#, \mathbbm{1}_a, \mathbbm{1}_{\#a>\#b}]_{tanh}
\label{eq:ht}
\end{equation}

This vector is (tanh-)one-hot in all cases except when $\#_a = \#_b$, in which case it zeros-out.

\subsection{Output layer}

The hidden vector $h_t$ is then multiplied by a linear layer $W_{out}$. We build the values of $W_{out}$ backwards based on the optimal target probabilities. 

Each \ANBN{} string has four phases: the start-of-sequence step, the `$a$' phase, the $b^{m<n}$ phase, and the final-`$b$' phase. Each row in the following matrix holds the optimal probabilities for the respective phase, based on PCFG (\ref{eq:pcfg-anbn}):

\begin{equation}
Targets = \begin{pmatrix}
p& 1 - p& 0 \\
0& 1 - p& p \\
0& 0& 1  \\
1& 0& 0 \\
\end{pmatrix}
\label{eq:probab-targets}
\end{equation}

Since the output layer feeds a final softmax, we build the logits backwards:

\[
W_{log} = ln(Targets + \varepsilon)
\]
with $\varepsilon$ preventing taking the log of 0. Here we use $\varepsilon=(2^{14}-1)^{-1}$.

Since we have four states and only three components in the hidden vector, we superimpose the four states onto three. First, split $W_{log}$ into $W_{out'}$ which contains the first three states, and a bias $b_{out}$ which contains the fourth:

\[
\begin{array}{l}
W_{out'} = {{W_{log}}_{1:3}} \\ 
b_{out} = {W_{log}}_{4}
\end{array}
\]

Then subtract to get:
\[
W_{out''} = (W_{out'} - b_{out})^T
\]

The transpose is taken so that multiplying by the one-hot $h_t$ copies columns.

Finally, divide by $tanh(1)$ because $h_t$ is tanh-one-hot based on (\ref{eq:ht}):

\[
    W_{out} = W_{out''} / tanh(1)
\]

As seen in (\ref{eq:ht}), $h_t$ is one-hot during all phases except the last `$b$', and thus copies the relevant probability distribution from (\ref{eq:probab-targets}). Adding the bias undoes the superimposition. $h_t$ is all-zero only when $\#_a=\#_b$, in which case $W_{out} \cdot h_t$ is zero. In this case the probabilities for the fourth state (final-$b$), stored in $b_{out}$, are outputted. 

\end{document}